\documentclass[letterpaper, 10 pt, conference]{ieeeconf}  %

\IEEEoverridecommandlockouts                              %

\usepackage{booktabs}       %
\usepackage{amsfonts}       %
\usepackage{amsmath,amssymb}
\usepackage{graphicx}
\usepackage{multirow}
\usepackage[export]{adjustbox}
\usepackage{xcolor}

\makeatletter
\let\NAT@parse\undefined
\makeatother

\usepackage[breaklinks=true,colorlinks,bookmarks=false]{hyperref}
\usepackage[capitalise]{cleveref}

\usepackage{url}

\renewcommand{\baselinestretch}{0.94}

\title{\LARGE \bf
Injecting Planning-Awareness into Prediction and Detection Evaluation
}

\author{Boris Ivanovic \hspace{1cm} Marco Pavone%
\thanks{*We thank Andrea Bajcsy, Karen Leung, Edward Schmerling, Shie Mannor, Nikolai Smolyanskiy and the rest of the NVIDIA AV Prediction team for many important discussions that shaped this work along the way.}%
\thanks{Boris Ivanovic is with NVIDIA Research {\{\tt\footnotesize bivanovic@nvidia.com\}}. Marco Pavone is with the Department of Aeronautics and
Astronautics, Stanford University, and with NVIDIA Research {\{\tt\footnotesize pavone@stanford.edu}, {\tt\footnotesize mpavone@nvidia.com}\}.}%
}

\begin{document}

\maketitle
\thispagestyle{empty}
\pagestyle{empty}

\begin{abstract}
Detecting other agents and forecasting their behavior is an integral part of the modern robotic autonomy stack, especially in safety-critical scenarios entailing human-robot interaction such as autonomous driving. Due to the importance of these components, there has been a significant amount of interest and research in perception and trajectory forecasting, resulting in a wide variety of approaches. Common to most works, however, is the use of the same few accuracy-based evaluation metrics, e.g., intersection-over-union, displacement error, log-likelihood, etc. While these metrics are informative, they are task-agnostic and outputs that are evaluated as equal can lead to vastly different outcomes in downstream planning and decision making. In this work, we take a step back and critically assess current evaluation metrics, proposing task-aware metrics as a better measure of performance in systems where they are deployed. Experiments on an illustrative simulation as well as real-world autonomous driving data validate that our proposed task-aware metrics are able to account for outcome asymmetry and provide a better estimate of a model's closed-loop performance.
\end{abstract}

\section{INTRODUCTION}
Detecting the presence of surrounding agents and predicting their future behavior is a necessary capability for modern robotic systems, especially as many autonomous systems are increasingly being deployed alongside humans in domains such as autonomous driving \cite{LefevreVasquezEtAl2014,BrouwerKloedenEtAl2016}, service robotics \cite{KrusePandeyEtAl2013,ChikYeongEtAl2016,LasotaFongEtAl2017}, and surveillance \cite{MorrisTrivedi2008,MurinoCristaniEtAl2017,HirakawaYamashitaEtAl2018}.
In particular, there has been a significant interest in object detection and trajectory forecasting within the autonomous driving community, with many major organizations incorporating state-of-the-art perception and behavior prediction algorithms within their vehicle software stack \cite{GMSafety2018,UberATGSafety2020,LyftSafety2020,WaymoSafety2021,ArgoSafety2021,MotionalSafety2021,ZooxSafety2021,NVIDIASafety2021}.
As a result, it is important to accurately evaluate the performance of detection and forecasting systems prior to their deployment. 

\begin{figure}[t]
    \centering
    \includegraphics[width=\linewidth]{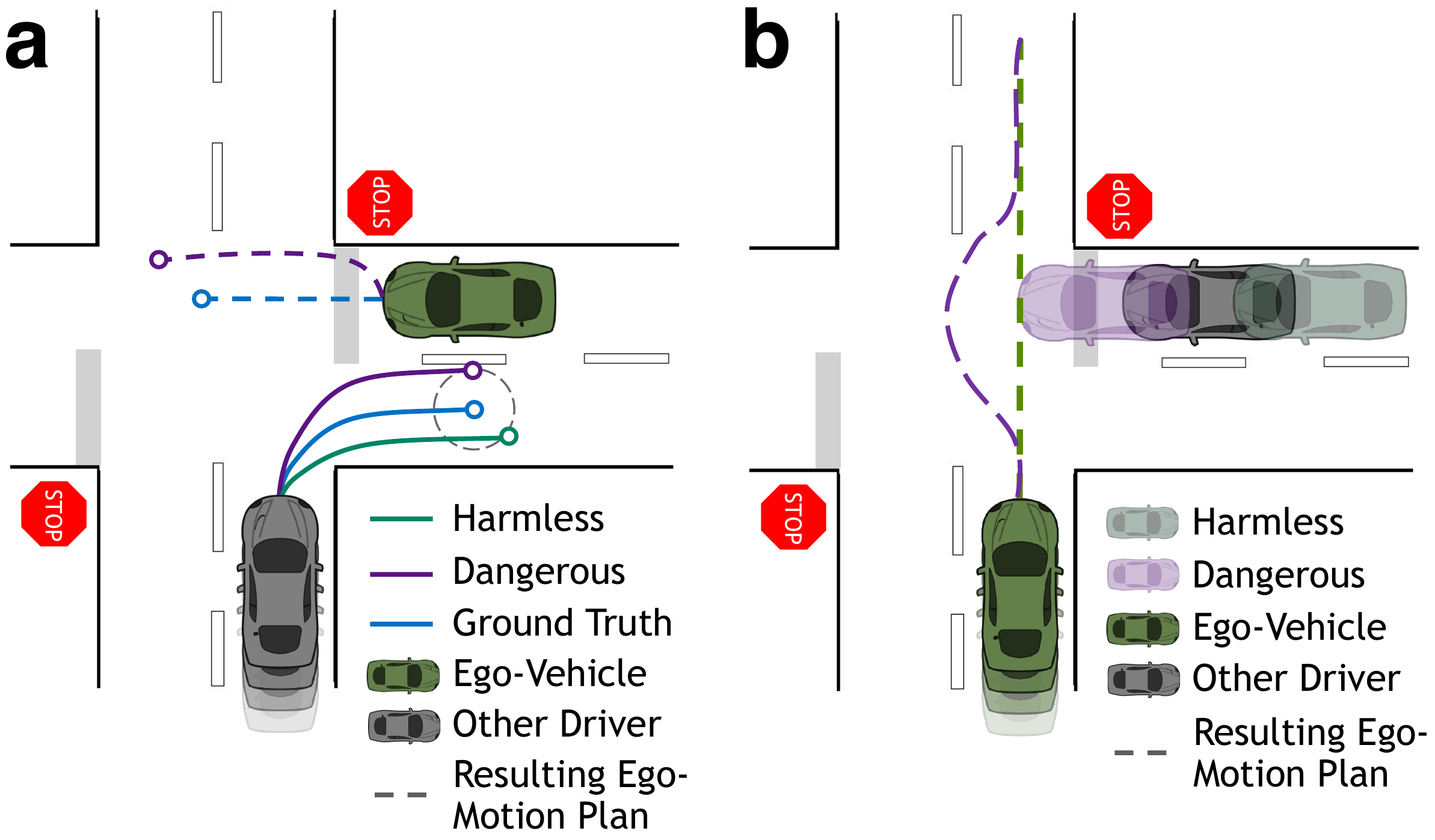}
    
    \vspace*{-0.25cm}
    
    \caption{\textbf{(a)} A human driver (gray) is about to turn right next to an autonomous vehicle (green). Two predictions (solid green and purple lines) of the human are also shown, which have the same metric accuracy.
    Although metrically equal, the purple prediction causes a safety-preserving maneuver (dashed purple) while the other does not affect the autonomous vehicle's motion plan (dashed blue). \textbf{(b)} In a similar scenario, an inaccurately-localized detection could make the ego-vehicle enact an unnecessary avoidance maneuver.}
    \label{fig:hero}
    
    \vspace*{-0.5cm}
    
\end{figure}

To date, nearly all works have relied on accuracy-based metrics such as intersection-over-union (IoU), average or final displacement error (ADE/FDE), negative log-likelihood (NLL), and other geometric or probabilistic quantities (see~\cite{PadillaNettoEtAl2020,PadillaPassosEtAl2021} and Table 1 of \cite{RudenkoPalmieriEtAl2019} for a comprehensive lists, as well as \cref{fig:metrics} for illustrated examples). At their core, accuracy-based metrics compare a model's predictions (e.g., a trajectory, bounding box, or distribution thereof) with ground truth (GT) values (e.g., a hand-labeled bounding box or future trajectory realized by an agent),
producing a value that quantifies how similar the two are. Comparing values solely based on similarity, however, does not consider downstream ramifications, and errant outputs with equal metric inaccuracy can lead to vastly different outcomes, examples of which are illustrated in \cref{fig:hero}.

{\bf Contributions.}
Towards this end, our contributions are twofold. First, we 
argue for the use of task-aware metrics to evaluate methods in a manner that better matches the systems in which they are deployed. Second, we present a novel augmentation to existing metrics that accounts for the effect that errant detections and predictions can have on an ego-vehicle's future motion plan. This is especially important for methods whose outputs are used to inform downstream planning and decision making, an arrangement commonly found in modern robotic autonomy stacks (e.g., \cite{WaymoSafety2021}). 

{\bf Organization.} The rest of the paper is organized as follows: \cref{sec:litreview} summarizes prior work on the evaluation of prediction and detection methods, identifying that the problem of task-aware evaluation has seldom been addressed thus far. \cref{sec:prob_form} rigorously formulates the problem tackled in this work and \cref{sec:method} describes our method for doing so. \cref{sec:expts} demonstrates the capabilities of our approach on an illustrative collision-avoidance simulation as well as real-world data. Lastly, \cref{sec:conclusion} concludes the work and highlights exciting potential future research directions. %

\section{RELATED WORK}\label{sec:litreview}

\begin{figure}[t]
    \centering
    \includegraphics[width=\linewidth]{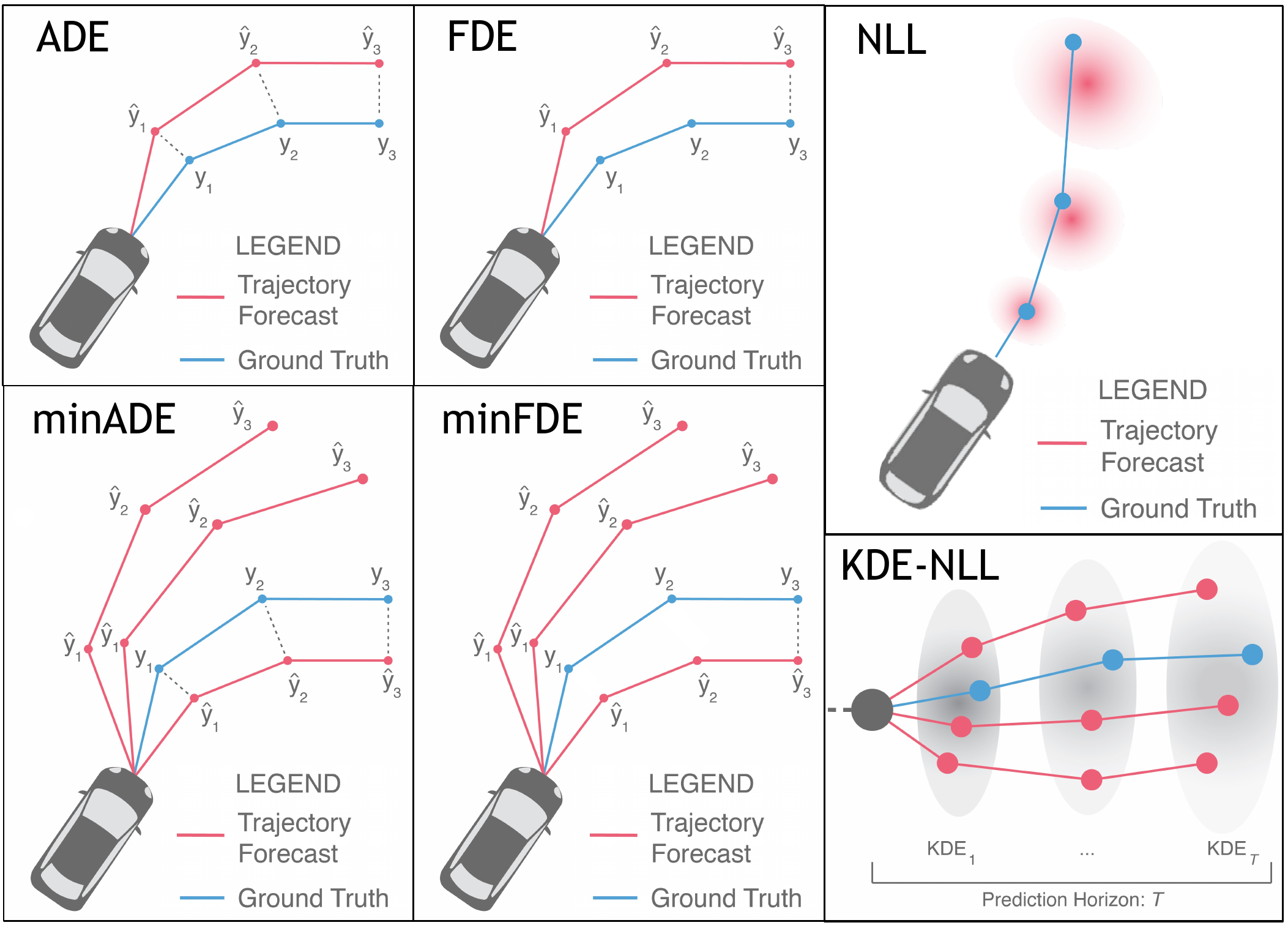}
    
    \vspace*{-0.25cm}
    
    \caption{Accuracy-based metrics broadly evaluate how similar a forecast trajectory (in red) is to the ground truth (in blue).}
    \label{fig:metrics}
    
    \vspace*{-0.5cm}
    
\end{figure}

{\bf Trajectory Forecasting Evaluation.} There has been a significant surge of interest in trajectory forecasting within the past decade, spawning a diverse set of approaches combining tools from physics, planning, and pattern recognition \cite{RudenkoPalmieriEtAl2019}. Accordingly, there have been many associated thrusts in developing prediction metrics that accurately evaluate these methods \cite{MorrisTrivedi2008,ZhangHuangEtAl2006,Zheng2015,QuehlHuEtAl2017}. Overall, two high-level classes of metrics have emerged: geometric and probabilistic. Geometric metrics (e.g.,~ADE and FDE) compare a single predicted trajectory to the GT, whereas probabilistic metrics (e.g., minimum ADE/FDE, NLL, kernel density estimate (KDE)-based NLL \cite{IvanovicPavone2019}) compare predicted distributions or sets of trajectories to the GT, taking into account additional information such as 
variance. A few examples of existing metrics are depicted in \cref{fig:metrics}.

While existing metrics are useful for evaluating the performance of trajectory forecasting methods in isolation, there are important 
considerations that arise during real-world deployment. Some examples include handling perception uncertainty in prediction \cite{LuoYangEtAl2018,DiHarakehEtAl2020,IvanovicLeeEtAl2021} and integrating prediction and planning \cite{KrusePandeyEtAl2013,SchmerlingLeungEtAl2018,NishimuraIvanovicEtAl2020,IvanovicElhafsiEtAl2020,SchaeferLeungEtAl2021}.
Most importantly, in this work we focus on the fact that
\emph{prediction errors are asymmetric in the real world}, i.e., predictions with the same metric accuracy may lead to vastly different outcomes,
an example of which is illustrated in \cref{fig:hero}~(a). Accounting for such asymmetry is of critical importance now that prediction algorithms are executed during planning in a closed-loop fashion (e.g., \cite{SchaeferLeungEtAl2021,TolstayaMahjourianEtAl2021}) and
deployed in real-world, safety-critical settings.

{\bf Object Detection Evaluation.} Object detection evaluation can be quite similar to trajectory forecasting evaluation. For example, the nuScenes detection challenge \cite{CaesarBankitiEtAl2019} utilizes metrics such as Average Translation/Orientation/Velocity Error (A\{TOV\}E), which are effectively the ADE of agent locations in the current frame as well as other state error values (heading and velocity). More generally, average precision (AP) is the most common metric used to evaluate object detectors \cite{PadillaNettoEtAl2020,PadillaPassosEtAl2021}. It computes the weighted mean of precisions achieved at each detection threshold, with the increase in recall from the previous threshold as the weight. To determine if an output is a ``correct" or ``incorrect" detection, its IoU is computed to the nearest GT object and compared to a set threshold (i.e., the detection threshold). Since IoU is positionally-symmetric (i.e., it does not matter \emph{where} the bounding box is, just its overlap with the GT), detection errors are also symmetric. However, \emph{detection errors are also asymmetric in the real world} and detections with the same IoU may lead to vastly different outcomes, an example of which is shown in \cref{fig:hero}~(b).

{\bf Task-Aware Metrics.} Comparatively, there are much fewer works on task-aware metrics for components of the autonomy stack (e.g., detection, tracking, prediction). One notable example
is the Planning KL Divergence (PKL) metric \cite{PhilionKarEtAl2020}. PKL measures how similar an object detector's outputs are to the GT by computing the difference between an ego-vehicle's plan using predicted detections and its plan using GT detections.
PKL compares the plan generated by a neural planner using GT detections as input (serving as the ``GT" plan) to the plan generated by the same planner using a given detector's output. This strategy directly evaluates detection algorithms by measuring their effects on downstream planning. 
A downside of this strategy is that it relies on the performance of a specific neural planner, which makes it difficult to determine if regressions in performance are caused by the detector or planner (e.g., could a better planner handle a wider range of detection errors? Do certain detectors only work well with certain planners?). Further, 
tools such as trajectory optimization, graph search, variational methods, and sampling-based motion planning are significantly more common than neural planning in real world systems \cite{PadenCapEtAl2016}.
In typical modern autonomous driving systems \cite{GMSafety2018,UberATGSafety2020,LyftSafety2020,WaymoSafety2021,ArgoSafety2021,MotionalSafety2021,ZooxSafety2021,NVIDIASafety2021}, a learned prediction model forecasts the future trajectories of agents around an ego-vehicle and a downstream planner uses the predictions to plan a smooth, collision-avoiding path to a goal state.
An intuitive initial idea for evaluating the predictor in this system is to compare the outputs of the planner when using the predicted trajectories versus the GT future trajectories, similar to the PKL metric~\cite{PhilionKarEtAl2020}. We will discuss the shortcomings of this strategy in \cref{sec:method}.

While not strictly about metrics, there are a few recent works which use prediction to better understand the interactivity of agents in the surrounding scene, either by directly incorporating interactivity in planning \cite{SchaeferLeungEtAl2021} or as a way to prioritize surrounding agents in planning \cite{TolstayaMahjourianEtAl2021}. 
In particular, Schaefer et al. \cite{SchaeferLeungEtAl2021} use a prediction model's gradients in trajectory optimization to produce ego-agent behaviors that minimally affect other agents. Similarly, Tolstaya et al. \cite{TolstayaMahjourianEtAl2021} train a prediction model which is then used to identify how much of an effect the ego-agent's actions have on other agents in the scene. %

\section{PROBLEM FORMULATION}\label{sec:prob_form}
We seek a function $f: O \times G \rightarrow \mathbb{R}$, where $O$ and $G$ respectively denote the method's output and GT space, that computes the accuracy of an object detection or trajectory forecasting method while accounting for real-world planning outcome asymmetry caused by errors (\cref{fig:hero}), e.g., by more heavily weighting dangerous errors. This definition is broad by design, there are many possible ways to achieve these characteristics. In the following, $\mathcal{A}$ denotes the set of agents in a scene, $\mathcal{A}'$ the set of detected objects, $K$ the number of prediction hypotheses (each with probability $p_k$), and $T$ the prediction horizon.

Without loss of generality, we take $O$ to be predicted bounding boxes (specified by $D_B$ values) $\mathbb{R}^{|\mathcal{A}'| \times D_B}$ for detection and predicted future trajectories $\mathbb{R}^{|\mathcal{A}| \times K \times T \times 2}$ with associated probabilities $\mathbb{R}^{|\mathcal{A}| \times K}$ for prediction. 
Other output formats such as predicted occupancy maps
can also be used (the following sections will discuss how to include them), but they are currently not as common as predicting multiple discrete trajectories. 
Similarly, we take $G$ to be bounding boxes $\mathbb{R}^{|\mathcal{A}| \times D_B}$ for detection and future trajectories $\mathbb{R}^{|\mathcal{A}| \times T \times 2}$ for prediction.

\section{TASK-AWARE EVALUATION METRICS}\label{sec:method}
In this section, we advocate for a set of core considerations that a task-aware metric should address and apply them to develop a planning-aware evaluation framework for prediction and detection.

Please note that we overload the term ``prediction" in the remainder of the paper to mean both detection and trajectory forecasting. We do so with the perspective that detection can be viewed as ``predicting" the locations of agents in the \emph{current} frame
whereas trajectory forecasting ``predicts" the locations of agents in \emph{future} frames.

\subsection{Core Desiderata for Task-Aware Metrics}
We advocate that task-aware evaluation metrics be:
\begin{enumerate}
    \item Able to capture asymmetries in downstream tasks.
    \item Task-aware and method-agnostic.
    \item Computationally feasible to compute.
    \item Interpretable.
\end{enumerate}
The first consideration directly addresses the core shortcoming of existing metrics, and is the main motivation for this work. In particular, it would be ideal to explicitly distinguish between harmless and catastrophic prediction errors, penalizing them in a similarly-asymmetric manner (i.e., weighting worse prediction outcomes more heavily).
The second desideratum is more nuanced, and requires task-aware metrics to be decoupled from any one particular method for that task. For example, a planning-aware prediction metric should not rely on a specific planner (e.g., FMT* \cite{JansonSchmerlingEtAl2015}, RRT* \cite{KaramanFrazzoli2011}) in its computation. The reason for this is that it harms applicability (not all practitioners use the same planner in their system of interest) and introduces biases (a specific planner may require a specific output format from an upstream prediction algorithm, which unfairly advantages methods that produce the same output format).
Importantly, these shortcomings apply directly to the intuitive predictor-planner evaluation idea described in \cref{sec:litreview}, making it arguably unsuitable for real-world use.
The third consideration ensures that practitioners are able to obtain metric values efficiently over modern large-scale datasets.
Finally, interpretability is important for any evaluation metric~\cite{Doshi-VelezKim2017}, allowing users to understand and contextualize the performance of their method.

\subsection{Injecting Planning-Awareness in Evaluation}
In this section, we outline a task-aware prediction metric that fulfills the aforementioned desiderata. In particular, we develop a planning-aware prediction metric for autonomous driving that evaluates trajectory forecasting and detection models based on their effect on an ego-vehicle's downstream motion planning. Specifically, our method identifies and heavily weights prediction errors that would be catastrophic.

At a high-level, our approach leverages human trajectory datasets (e.g., \cite{CaesarBankitiEtAl2019,HoustonZuidhofEtAl2020,SunKretzschmarEtAl2020}) to learn a planning cost function whose sensitivities to prediction outputs determine which agents most influence planning. These sensitivities can then be used to inject task-awareness within existing metrics (e.g., by weighting prediction errors based on their planning influence).

\begin{figure*}[t]
    \centering
    \includegraphics[width=\linewidth]{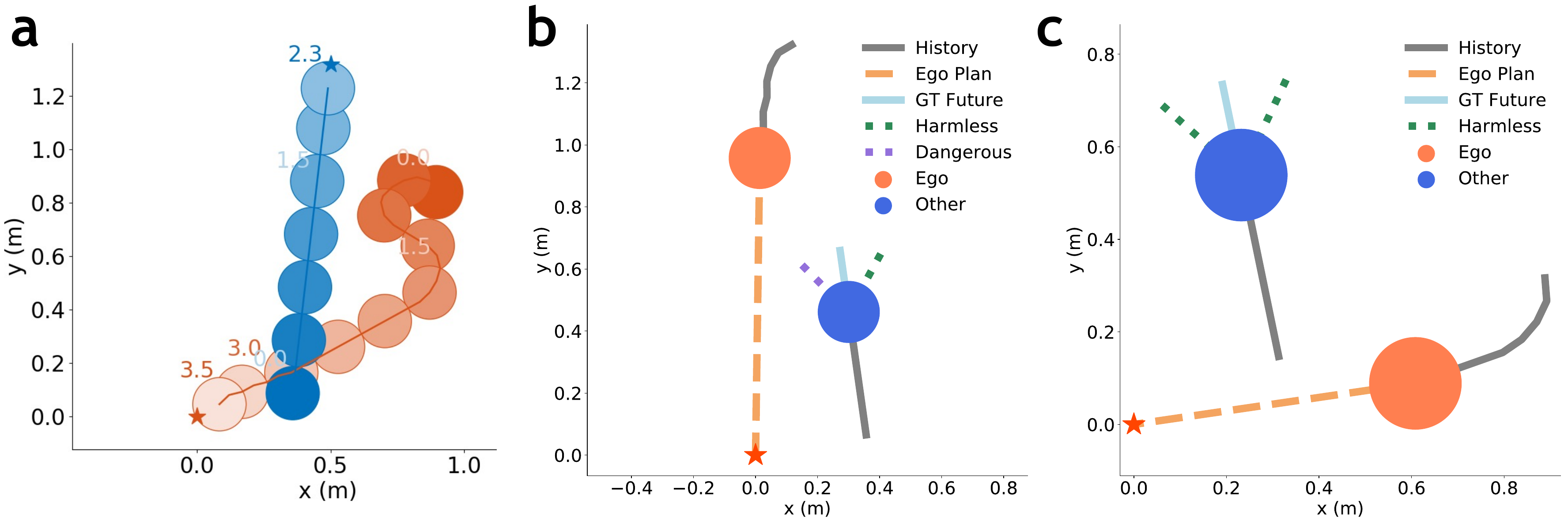}
    
    \vspace*{-0.25cm}
    
    \caption{
    \textbf{(a)}. An ego-vehicle (orange) maneuvers to the origin while avoiding another agent (blue), lighter colors occur later. 
    Importantly, our method is able to distinguish between metrically-equal errant predictions in a planning-aware manner. In a head-on scenario \textbf{(b)}, planning sensitivities are much higher for predictions that veer into the ego-vehicle's path (purple dashed) compared to those that steer away (green dashed). Further, when it is unlikely that an agent would influence the ego-vehicle's plan \textbf{(c)}, our method yields small planning sensitivities for all predictions.}
    \label{fig:expt}
    
    \vspace*{-0.5cm}
    
\end{figure*}

{\bf Planning Cost Functions.}
Since using a specific planner would go against the second desideratum, we instead work with planning \emph{cost functions}. Cost functions are general (defining the high-level goal for a planning problem) and planner-agnostic (the same cost function can be minimized with different planners, leading to different output paths). Which cost function to use is an important consideration, and we maintain method-agnosticism by 
learning a proxy cost function whose minimization well-reproduces ego-vehicle trajectories in an autonomous driving dataset. Importantly,
we use datasets with a human-driven ego-vehicle to maintain planner-agnosticism.
In particular, given the $D_S$-dimensional dynamic states $\mathbf{s}^{(t)} \in \mathbb{R}^{(|\mathcal{A}|+1) D_S}$ (e.g., position, velocity, acceleration) of an ego-vehicle and other agents $a \in \mathcal{A}$ in the scene, the $D_U$-dimensional actions enacted by the ego-vehicle $\mathbf{u}_\text{R}^{(t)} \in \mathbb{R}^{D_U}$, as well as $D_P$-dimensional predictions of other agents' future positions for the next $T$ timesteps $\mathbf{\hat{s}}^{(t:T)} \in \mathbb{R}^{T \times |\mathcal{A}| D_P}$, we specify a linear cost function $c$ of the form
$c(\mathbf{s}^{(t)}, \mathbf{u}_\text{R}^{(t)}, \mathbf{\hat{s}}^{(t:T)}) = \theta^T \phi(\mathbf{s}^{(t)}, \mathbf{u}_\text{R}^{(t)}, \mathbf{\hat{s}}^{(t:T)})$,
where $\theta_i \in \mathbb{R}$ is the weight of the $i^\text{th}$ feature $\phi_i: \mathbb{R}^{(|\mathcal{A}|+1) D_S \times D_U} \rightarrow \mathbb{R}$. 
Our motivation for using a linear function is that it additively combines features that are considered by the ego-vehicle during planning (leading to easier interpretation \cite{Molnar2019LinReg}), is efficient to compute, and, although simple, we will show in \cref{sec:expts} that it is already useful for planning-aware
evaluation.

{\bf Planning Sensitivity for Metric Weighting.} To learn the weights $\theta_i$, we leverage Continuous Inverse Optimal Control (CIOC)~\cite{LevineKoltun2012}. Importantly, CIOC does not require globally-optimal expert trajectories in general~\cite{LevineKoltun2012}, thus we only assume that human trajectories are \emph{locally} optimal. 
With the learned cost function in hand, we compute its sensitivity with respect to the predicted agent locations by obtaining the 
gradient $\nabla_{\mathbf{\hat{s}}^{(t:T)}} c$. The gradient magnitudes are a measure of how sensitive the ego-vehicle's plan is to each agent's prediction, with higher values indicating more sensitivity. Finally, these values can be used to weight predictions
within existing accuracy-based metrics. For example, planning-informed (PI) versions of accuracy-based metrics can be implemented
as:
\begin{equation}\label{eqn:pimetric}
    \text{PI-Metric} = \frac{1}{|\mathcal{A}|} \sum_{a \in \mathcal{A}} f(a, |\nabla_{\mathbf{\hat{s}}^{(t:T)}} c|) \cdot  \text{Metric}(\mathbf{\hat{s}}^{(t:T)}_a, \mathbf{s}^{(t:T)}_a),
\end{equation}
where ``Metric" is a placeholder for existing metrics (e.g., those in \cref{fig:metrics}) and $f$ allows for the implementation of different weighting schemes, e.g., normalization (where $f(a, g) = 1 + g_a / \sum_{a' \in \mathcal{A}} g_{a'}$) or softmax (where $f(a, g) = 1 + \exp(g_a) / \sum_{a' \in \mathcal{A}} \exp(g_{a'})$).

\section{EXPERIMENTS}\label{sec:expts}

With a way to obtain planning sensitivities and weighted metrics, we now demonstrate the capabilities of our proposed planning-aware metric\footnote{Code can be found at \url{https://github.com/BorisIvanovic/PlanningAwareEvaluation}.} in an illustrative collision-avoidance scenario as well as on real-world human driving data.

\subsection{Illustrative Collision-Avoidance Scenario}\label{sec:particles}
We first illustrate the performance of our proposed metric augmentation in a controlled simulation environment where an ego-vehicle with unicycle dynamics~\cite{LaValle2006Unicycle} is tasked with reaching the origin from a random starting state while avoiding collisions with surrounding vehicles. An example rollout is shown in \cref{fig:expt} (a).

{\bf Planning Cost Function.}
Our cost function contains four terms: A goal term (squared distance between the ego-vehicle and the origin), a control term (squared magnitude of the control effort), and two collision-avoidance terms which are Gaussian radial basis functions (RBFs) \cite{Powell1987}, $\varphi(\cdot)$, centered at the other agent's current and one-step predicted positions. Formally,
\begin{equation}\label{eqn:toy_cost}
\begin{aligned}
    c(&\mathbf{s}^{(t)}, \mathbf{u}_\text{R}^{(t)}, \mathbf{\hat{s}}^{(t:T)}) = \theta_1 \|\mathbf{s}^{(t)}_{\text{ego}}\|_2^2 + \theta_2 \|\mathbf{u}_\text{R}^{(t)}\|_2^2\\
    &+ \theta_3 \varphi(\|\mathbf{s}^{(t)}_{\text{ego}} - \mathbf{s}^{(t)}_{a}\|) + \theta_4 \varphi(\|\mathbf{\hat{s}}^{(t+1)}_{\text{ego}} - \mathbf{\hat{s}}^{(t+1)}_{a}\|),
\end{aligned}
\end{equation}
where $\mathbf{s}^{(t)}_{\text{ego}}, \mathbf{s}^{(t)}_{a} \in \mathbb{R}^2$ are the $x,y$ positions of the ego-vehicle and other agent, and $\mathbf{s}^{(t+1)}_{\text{ego}}, \mathbf{s}^{(t+1)}_{a}\in \mathbb{R}^2$ are the one-step predicted positions of the ego-vehicle and other agent (i.e., one-step constant-velocity predictions).%

We chose these terms as they are the minimal set of terms which encourage progress towards a goal, control parsimony, reactive collision avoidance (via detections in the current frame), and proactive collision avoidance (by avoiding where agents are predicted to be in the future). RBFs, in particular, indicate that collisions are undesirable by producing (non-convex) regions of increased cost in an otherwise purely-quadratic function. In the case of different model output formats (e.g., occupancy maps), the last two terms would change, e.g., to the estimated likelihood of a collision. %

{\bf Learned Weights.} To learn $\theta$, 64 rollouts\footnote{We experimented with different numbers of rollouts, but any more than 64 yielded nearly identical $\theta$ values (intuitive as there are only 4 parameters to learn) at the cost of increased compute times.} of a pre-trained GPU/CPU Asynchronous Advantage Actor-Critic for Collision Avoidance with Deep RL (GA3C-CADRL) \cite{EverettChenEtAl2021} policy were collected in an associated collision avoidance Gym environment \cite{EverettChenEtAl2021} (\cref{fig:expt}~(a) shows one such rollout).
The cost function term weights, $\theta$, were then obtained from the rollouts with CIOC via maximum likelihood estimation (using PyTorch's L-BFGS \cite{ByrdLuEtAl1995} optimizer)~\cite{LevineKoltun2012}, yielding $\theta = [1.21, 4.19, 0.37, 0.35]$. These $\theta_i$ are sensible as all are positive and align with GA3C-CADRL's behavior, i.e., it minimally changes its heading and moves at a constant velocity (high $\theta_2$) towards the goal (moderate $\theta_1$). Most interestingly, GA3C-CADRL was trained with a goal reward that is $4\times$ higher than its collision penalty \cite{EverettChenEtAl2021}, which is almost exactly what our method recovers ($\theta_1$ vs.~$\theta_3,\theta_4$).
Finally, the ego-vehicle's planning sensitivity to agent predictions, $\nabla_{\mathbf{\hat{s}}^{(t:T)}} c$, is obtained with standard PyTorch autodifferentiation tools~\cite{PaszkeGrossEtAl2017}. 

{\bf Planning-Aware Prediction Evaluation.}
We analyze our method's performance in two scenarios. \cref{fig:expt} (b) shows a scene where the ego-vehicle must avoid a head-on collision with another agent to reach its goal. Two predictions with the exact same raw displacement errors (ADE = $0.075$, FDE = $0.15$) are also depicted (dashed purple and green). The resulting gradient magnitudes are $0.90$ (purple) and $0.21$ (green), sensibly indicating that the purple prediction affects the ego-vehicle's planning the most (it veers directly into the ego-vehicle's desired path). Applying \cref{eqn:pimetric} with these values yields a $25\%$ higher piADE and piFDE for the purple prediction than the green prediction, with $f(a, g) = 1 + \max(0, g_a - g_{a_{GT}})$ where $g_{a}$ and $g_{a_{GT}} = 0.57$ are the planning sensitivities of the prediction and GT future. 

In \cref{fig:expt} (c), we sanity check the performance of our method when there is little planning influence from another agent. As expected, the planning sensitivities are all near zero, yielding piADE and piFDE that match their task-agnostic counterparts ADE and FDE.

These results are encouraging as they demonstrate that our method can identify prediction outcome asymmetries and
incorporate such information within existing metrics, 
providing a measure of downstream task performance. 

\begin{figure}[t]
    \centering
    \includegraphics[width=\linewidth]{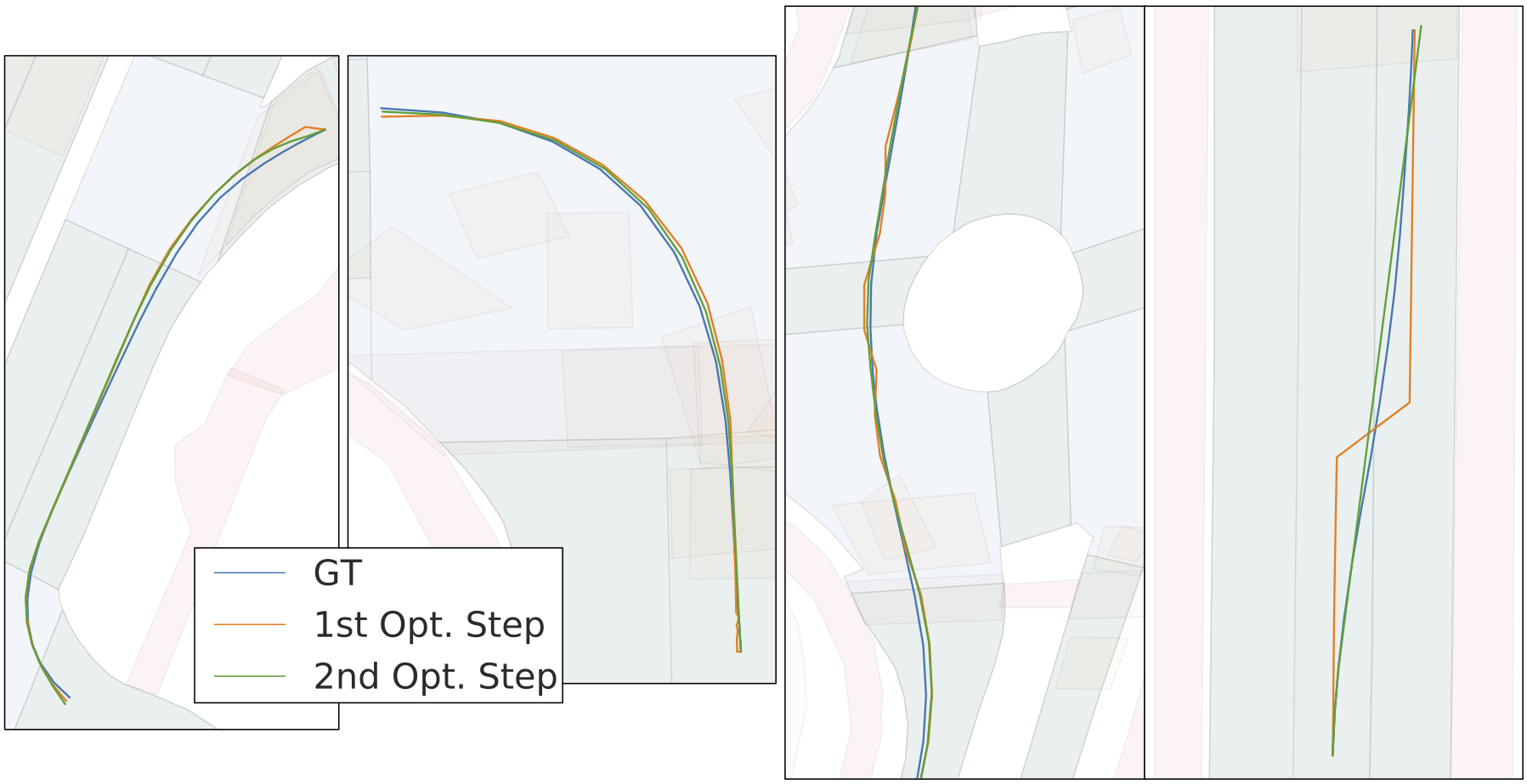}
    
    \vspace*{-0.25cm}
    
    \caption{To evaluate the planning cost function and learned weights, we optimize \cref{eqn:nuScenes_cost} and compare the resulting trajectories with the ground truth (GT) ego-vehicle motion on the validation set. Our method's resulting trajectories closely match the GT, even in scenarios with turns across multiple roads, roundabouts with noisy lane annotations, and lane changes.}
    \label{fig:reopt}
    
    \vspace*{-0.25cm}
    
\end{figure}

\begin{figure}[t]
    \centering
    \includegraphics[width=\linewidth]{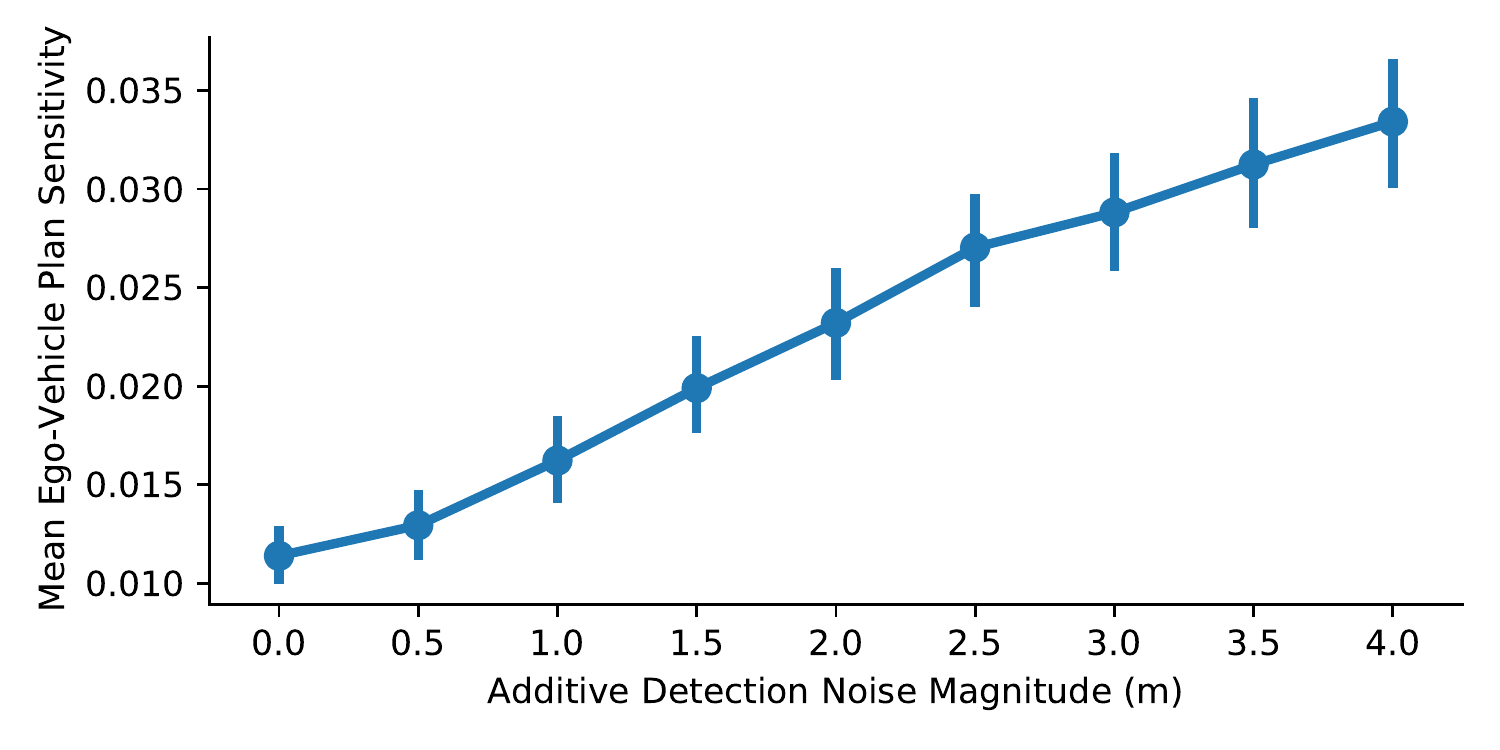}
    
    \vspace*{-0.25cm}
    
    \caption{To simulate the behavior of our planning sensitivity scheme under detection error, we added noise to the ground truth agent detections. As can be seen, worse (noisier) detections lead to worse (higher) ego-vehicle planning sensitivities. Error bars are 95\% confidence intervals.}
    \label{fig:noisy_dets}
    
    \vspace*{-0.5cm}
    
\end{figure}

\begin{figure}[t]
    \centering
    \includegraphics[width=0.49\linewidth]{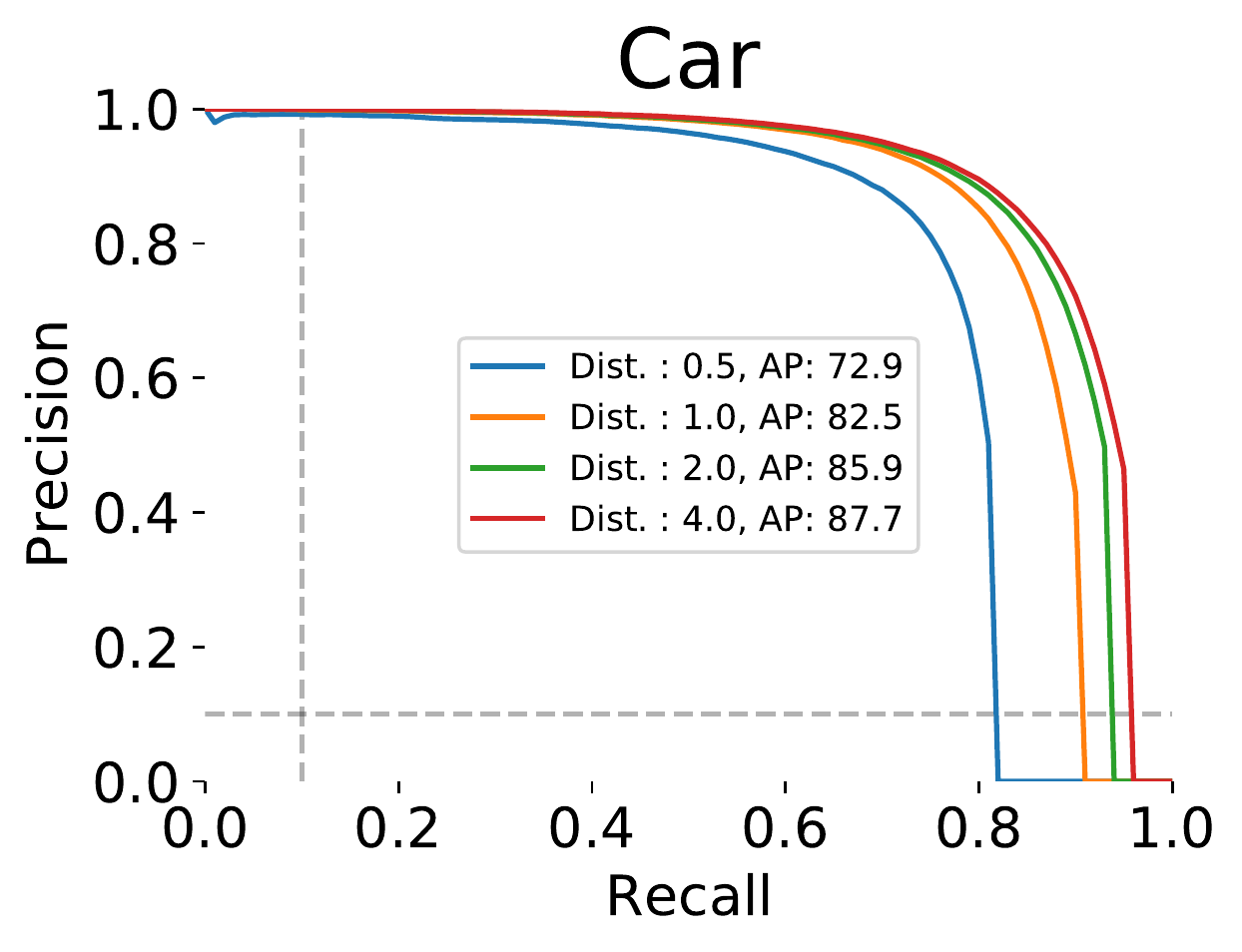}
    \includegraphics[width=0.49\linewidth]{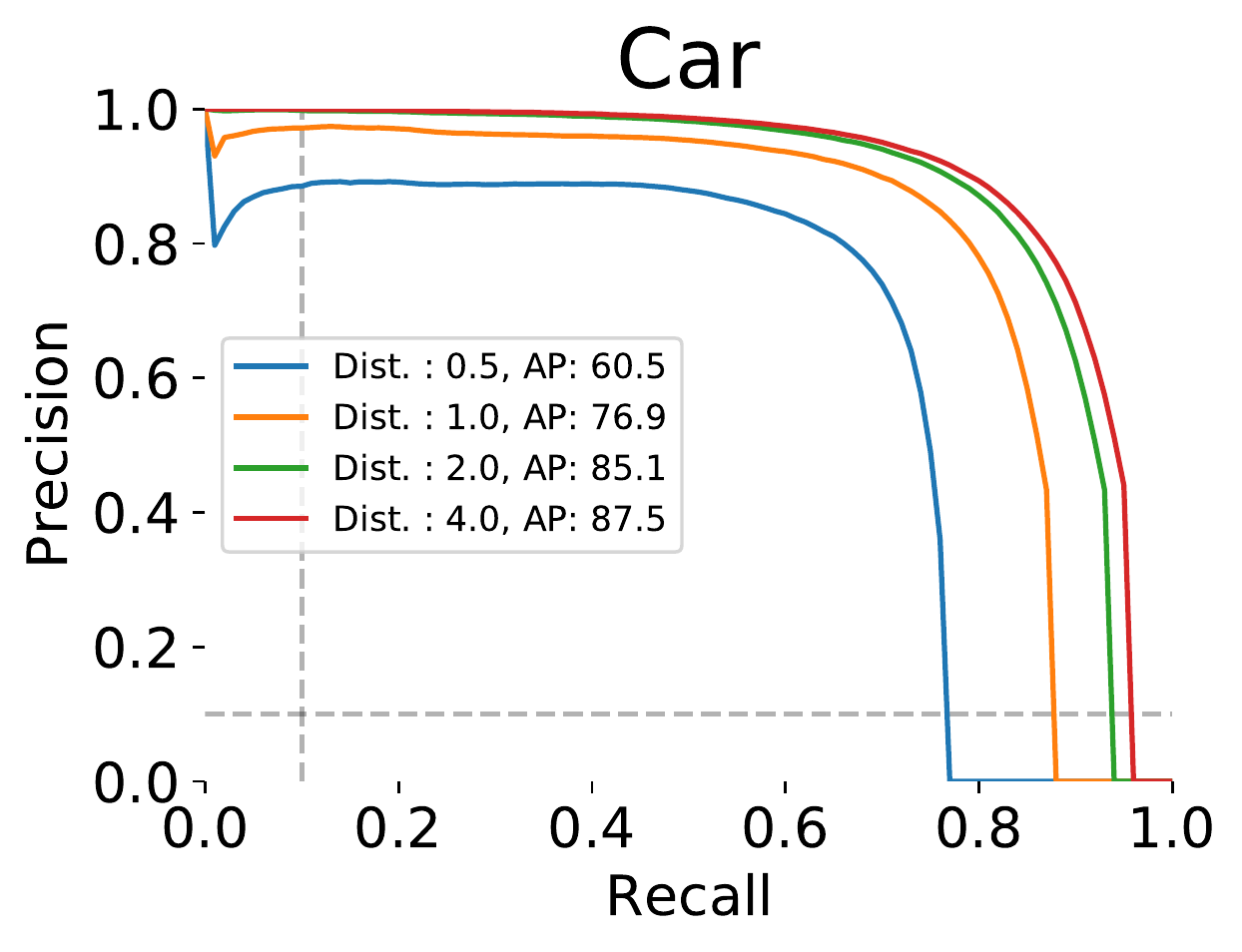}

    \vspace*{-0.25cm}
    
    \caption{To inject planning-awareness into detection metrics, we scale the nuScenes detection threshold based on the ego-vehicle's planning sensitivity
    so that highly planning-sensitive agents must be detected more accurately to register as true positives.
    \textbf{Left}: Original PR curves and AP values for the Megvii detector~\cite{MEGVII2019} on nuScenes~\cite{CaesarBankitiEtAl2019} cars. \textbf{Right}: The same, after introducing our sensitivity-based threshold scaling scheme.}
    \label{fig:det_reweight_quant}
    
    \vspace*{-0.25cm}
    
\end{figure}

\begin{figure}[t]
    \centering
    \includegraphics[width=.48\linewidth]{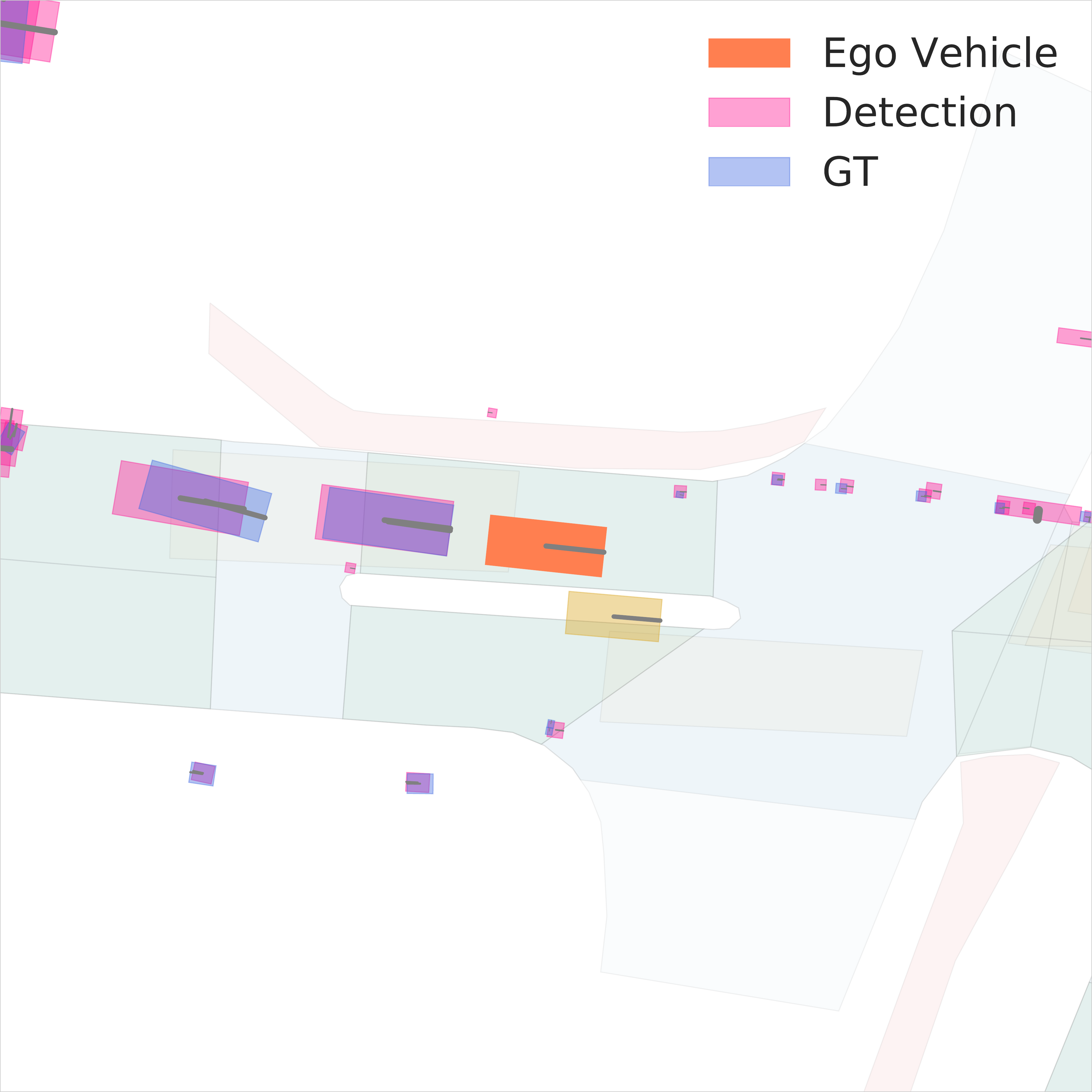}
    \includegraphics[width=.48\linewidth]{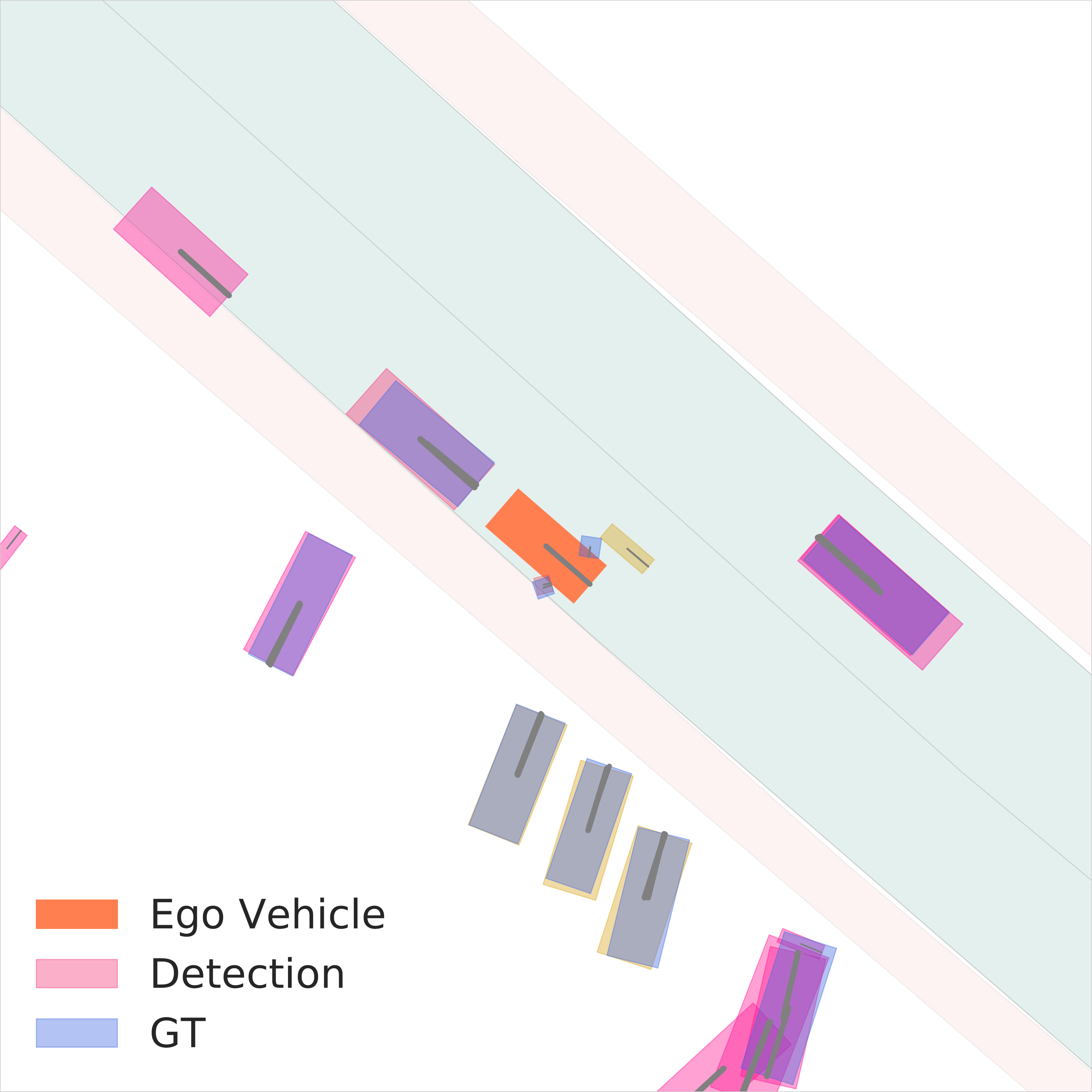}
    
    \vspace*{-0.25cm}
    
    \caption{Our method is able to identify detections to which the ego-vehicle's planning is highly sensitive (in gold), including nearby false positives (\textbf{left}) and agents which may enter the road near the ego-vehicle (\textbf{right}). Accurate detections are marked in purple (mixing blue GT and red detections).
    }
    \label{fig:det_qual}
    
    \vspace*{-0.5cm}
    
\end{figure}

\begin{figure}[t]
    \centering
    \includegraphics[width=0.49\linewidth]{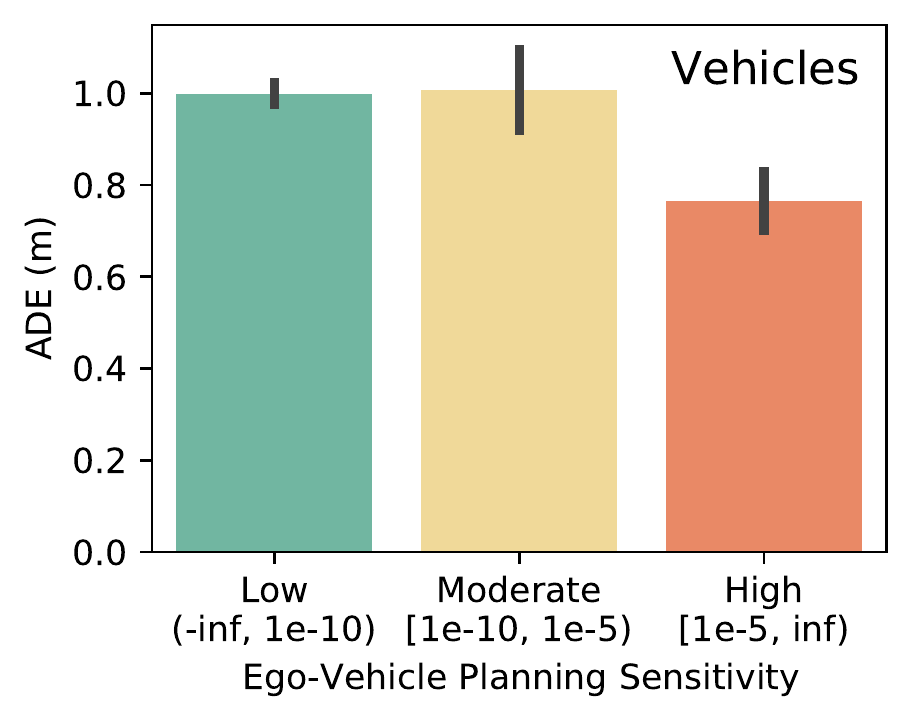}
    \includegraphics[width=0.49\linewidth]{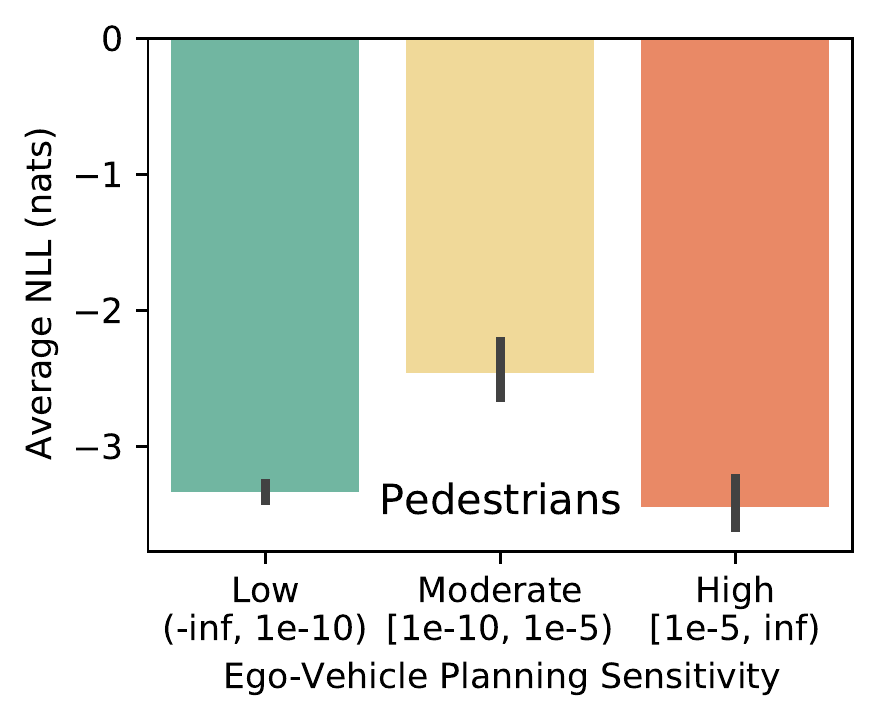}

    \vspace*{-0.25cm}
    
    \caption{Our method can striate errors made by detection models depending on the ego-vehicle's planning sensitivity to the agent being predicted.
    \textbf{Left}: Encouragingly, Trajectron++ \cite{SalzmannIvanovicEtAl2020}'s most-likely predictions are most accurate for vehicles that heavily influence planning. \textbf{Right}: Its full distributional output is equally accurate for pedestrians that lightly and heavily influence planning, with performance regressions on pedestrians with moderate planning sensitivities. Error bars are 95\% confidence intervals.}
    \label{fig:pred_reweight_quant}
    
    \vspace*{-0.5cm}
    
\end{figure}

\subsection{Autonomous Driving}
To assess the performance of our proposed task-aware metric in real-world settings, we evaluate existing state-of-the-art methods for prediction~\cite{SalzmannIvanovicEtAl2020} and detection~\cite{MEGVII2019} on 
the large-scale nuScenes dataset~\cite{CaesarBankitiEtAl2019}. nuScenes is comprised of 1000 autonomous driving scenes in Boston and Singapore, each of which are 20s long and annotated at 2 Hz ($\Delta t = 0.5s$), with up to 23 semantic object classes. Similar to \cref{sec:particles}, the ego-vehicle is modeled as a dynamically-extended unicycle \cite{LaValle2006BetterUnicycle}. We use 128 scenes from the nuScenes prediction challenge training set to learn $\theta$
and evaluate on the provided validation set. 

{\bf Planning Cost Function.}
Our cost function contains six terms, accounting for lateral lane deviation, %
lane orientation deviation, %
control effort, %
and two collision-avoidance terms (similar to \cref{sec:particles}).
Formally,
\begin{align}\label{eqn:nuScenes_cost}
    c(&\mathbf{s}^{(t)}, \mathbf{u}_\text{R}^{(t)}, \mathbf{\hat{s}}^{(t:T)}) = \theta_1 \|\mathbf{s}^{(t)}_{\text{ego}} - \mathbf{s}_{\ell,\perp}^{(t)}\|_2^2 + \theta_2 (\psi^{(t)}_{\text{ego}} - \psi^{(t)}_{\ell,\perp})^2 \nonumber\\
    &+ \theta_3 \|\mathbf{s}^{(t)}_{\text{ego}} - \mathbf{s}_{g}\|_2^2 + \theta_4 \varphi\left(\min_{a \in \mathcal{A}} \|\mathbf{s}^{(t)}_{\text{ego}} - \mathbf{s}^{(t)}_{a}\|\right) + \theta_5 \|\mathbf{u}_\text{R}^{(t)}\|_2^2 \nonumber\\
    &+ \theta_6 \varphi\left(\min_{a \in \mathcal{A}}\ \sum_{k=1}^K p_k \left[\min_{t' \in \{1, T\}}\|\mathbf{\hat{s}}^{(t + t')}_{\text{ego}} - \mathbf{\hat{s}}^{(t + t')}_{a, k}\|\right]\right)
\end{align}
where $\ell$ is the lane closest to the ego-vehicle, $\mathbf{s}^{(t)}_{\text{ego}}, \mathbf{s}_g, \mathbf{s}^{(t)}_{a}, \mathbf{s}^{(t)}_{\ell,\perp} \in \mathbb{R}^2$ are the $x,y$ positions of the ego-vehicle, its goal, agent $a \in \mathcal{A}$, and closest point to the ego-vehicle from lane $\ell$'s centerline, and $\psi^{(t)}_{\text{ego}}, \psi^{(t)}_{\ell,\perp} \in \mathbb{R}$ are the heading of the ego-vehicle and its closest point on lane $\ell$'s centerline.

The $\theta_3, \theta_5$ terms in \cref{eqn:nuScenes_cost} were chosen as in \cref{sec:particles} (to capture progress and control parsimony). Lane information ($\theta_1, \theta_2$) provides important inductive biases (e.g., that the ego-vehicle's overall motion can be subdivided into a collection of local effects, such as lane center deviation) that make the problem of learning $\theta$ significantly more tractable.
Terms $\theta_4, \theta_6$ capture collision avoidance by computing the closest that any agent $a \in \mathcal{A}$ is to the ego-vehicle ($\theta_4$) and the expected closest distance another agent will be to the ego-vehicle in the next $T$ timesteps ($\theta_6$).

{\bf Learned Weights.} In order to maintain method-agnosticism, we use the GT positions of agents for $\mathbf{s}_a^{(t)}$ (rather than a specific detector) and the GT future positions of agents for $\mathbf{\hat{s}}^{(t:T)}$ (rather than a specific predictor) to learn~$\theta$. We set $T = 3$s in order to match the prediction horizon commonly used by existing approaches on the nuScenes dataset \cite{SalzmannIvanovicEtAl2020}.
Training with CIOC yields $\theta=[1.722, 0.562, 3\times 10^{-6}, 11.865, 1.352, 0.241]$, which is sensible as all values are positive and most are of similar scale. Interestingly, the reactive collision avoidance term $\theta_4$ is quite large, indicating that the ego-vehicle strongly prefers to be far away from other agents. Additionally, the goal term $\theta_3$ is very small since the squared distances between the ego-vehicle and its final position can be quite large, thus CIOC compensates by accordingly adjusting the scale of $\theta_3$.

{\bf Plan Reoptimization.} To evaluate if our planning cost function and learned weights $\theta$ can reproduce human driving behavior, we minimize \cref{eqn:nuScenes_cost} for each test scene and compute the difference between the resulting ego-vehicle trajectories and the GT ego-vehicle trajectories. Since \cref{eqn:nuScenes_cost} is non-convex, we optimize it with a two-step approach: First, the convex parts of \cref{eqn:nuScenes_cost} (i.e., $\theta_1$, $\theta_2$, $\theta_3$, and $\theta_5$) are minimized\footnote{We use OSQP \cite{StellatoBanjacEtAl2020} to solve the resulting quadratic problem.} to obtain a candidate ego-vehicle state and control trajectory. The states and controls are then used as an initial guess when minimizing the full cost function with a non-convex solver\footnote{We used SciPy's \texttt{optimize.minimize} function.}. \cref{fig:reopt} visualizes a few resulting trajectories, all of which are nearly identical to the GT trajectory. Quantitatively, in the case of detection evaluation (i.e., dropping the prediction cost term), the average maximum $x, y$ errors (i.e., the mean over all scenes of the maximum $x,y$ error per scene) are $\{0.627, 0.696\}$m, respectively. Including the prediction cost term (as if evaluating prediction) yields average maximum $x,y$ errors of $\{0.585, 0.661\}$m, indicating that using more information (i.e., predictions) yields better ego-vehicle trajectories. Overall, both configurations of \cref{eqn:nuScenes_cost} yield accurate optimized trajectories, indicating that our planning cost function and learned weights $\theta$ can faithfully reproduce human driving behavior.

{\bf Planning-Awareness in Detection.} 
To understand the effect of errant detections on our proposed metric, we compute the ego-vehicle's planning sensitivity to increasingly erroneous detections
by adding zero-mean Gaussian noise with increasing variance to GT agent positions. \cref{fig:noisy_dets} shows that there is a sensible and interpretable linear relationship between the magnitude of the added noise and resulting ego-vehicle planning sensitivity. Intuitively, worse detections are more likely to affect motion planning. %

To demonstrate how our metric can be used to evaluate a detector, we assess the open-source Megvii detector~\cite{MEGVII2019} on the nuScenes validation set. We inject planning-awareness into AP computation by scaling the nuScenes detection threshold\footnote{Specifically, Line 88 of \url{https://github.com/nutonomy/nuscenes-devkit/blob/master/python-sdk/nuscenes/eval/detection/algo.py}.} according to the ego-vehicle's planning sensitivity to the agent being detected. In particular, we divide the detection threshold by $(1 + |\nabla_{\mathbf{s}^{(t)}_a} c|)$, requiring detectors to be more accurate on highly-sensitive agents to register true positives. Since this only maintains or reduces the detection threshold, the best result would be achieving the same performance before threshold-scaling (meaning that all detection errors were made on agents with low planning sensitivity). As shown in \cref{fig:det_reweight_quant}, introducing this threshold-scaling scheme worsens performance, indicating that the Megvii detector makes errors which are likely to affect motion planning.

Qualitatively, the Megvii detector's most sensitive detection errors are visualized in  \cref{fig:det_qual}. Intuitively, they are false positive detections which are immediately next to the ego-vehicle, significantly affecting motion planning in that frame.

\begin{figure}[t]
    \centering
    \includegraphics[width=\linewidth]{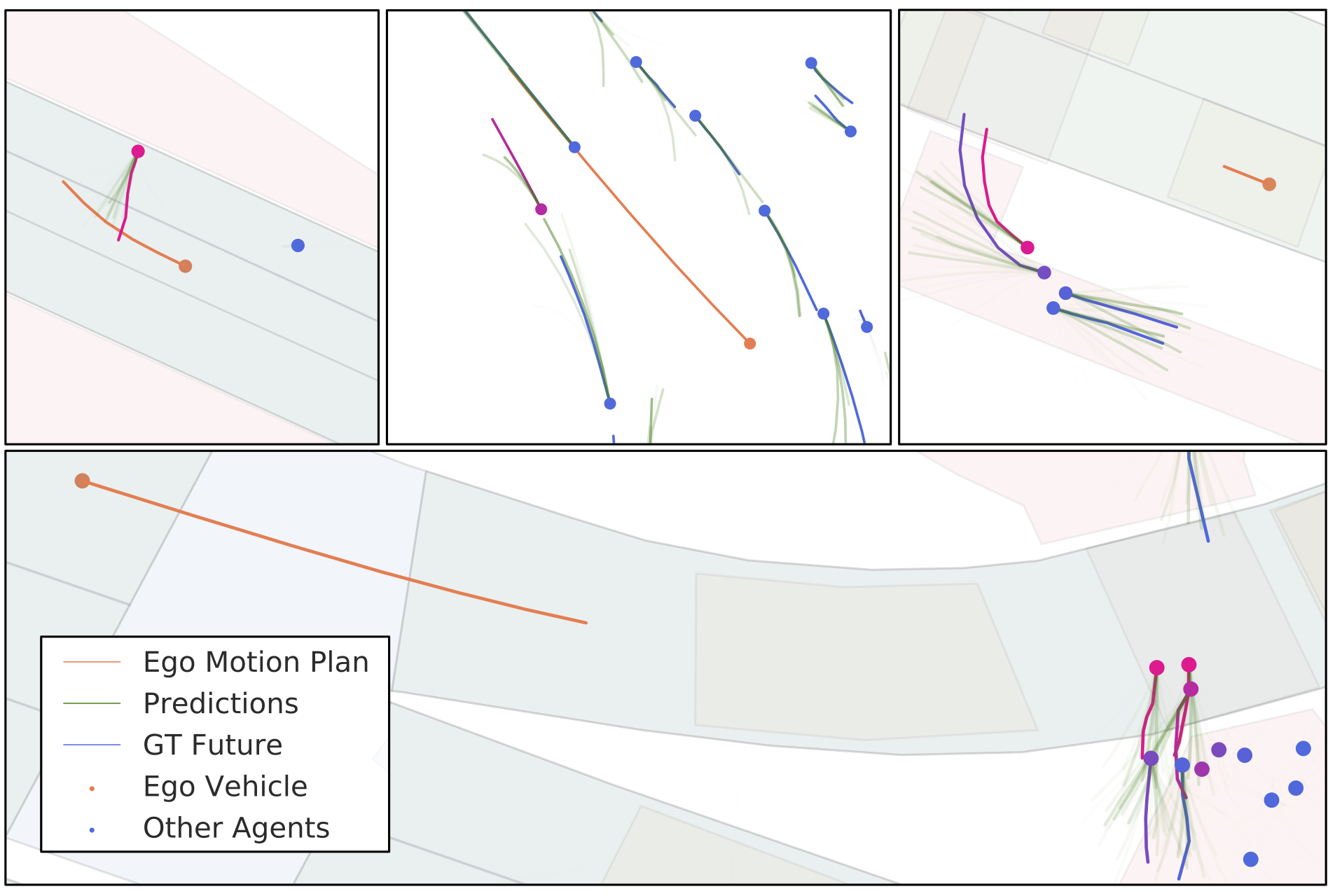}

    \vspace*{-0.25cm}
    
    \caption{Our method is able to identify which agents most influence the ego-vehicle's motion planning (in red or purple), weighting their prediction errors more heavily than an ineffectual agent's prediction errors. Our method highlights scenarios
    such as a group of pedestrians crossing a street, a pedestrian jay-walking, a vehicle merging, and pedestrians about to cross at a cross-walk, all of which cause the ego-vehicle to either reduce its speed, turn, or stop.}
    \label{fig:pred_qual}
    
    \vspace*{-0.5cm}
    
\end{figure}

{\bf Planning-Awareness in Prediction.} Finally, we assess the state-of-the-art trajectory forecasting method Trajectron++ \cite{SalzmannIvanovicEtAl2020} on the nuScenes validation set in a planning-aware fashion. In particular, we compute the model's prediction errors with both deterministic (e.g., ADE) and probabilistic (e.g., NLL) metrics, striating them by the predicted agent's planning sensitivity. \cref{fig:pred_reweight_quant} visualizes for which agents Trajectron++ performs the worst (namely, those which moderately affect motion planning). Encouragingly, the model performs best on highly planning-sensitive agents, indicating that Trajectron++ produces its best predictions in scenarios where predictions are important (i.e., for planning).

Qualitatively, our method's planning sensitivities capture many interesting scenarios (a few of which are shown in \cref{fig:pred_qual}), producing high planning sensitivities for agents which ultimately affect the ego-vehicle's motion (e.g., by making it turn or stop). \cref{fig:pred_qual} also shows why heuristics such as distance are insufficient proxies for agent importance or planning sensitivity, since the agents closest to the ego-vehicle are usually ineffectual (e.g., pedestrians on sidewalks).

\section{CONCLUSION}\label{sec:conclusion}

In this work, we advocate for the incorporation of task-awareness in detection and trajectory forecasting evaluation. In particular, we show by way of example that existing metrics neglect the asymmetries of real-world prediction outcomes, outline four core considerations that any task-aware metric should address, and provide a framework that injects task-awareness into existing prediction and detection metrics. Looking forward, this work further serves as a call to action for additional research in task-aware evaluation metrics, especially now that the field has reached a level of maturity where state-of-the-art methods are commonly deployed in real-world, safety-critical settings. Evaluation methods should reflect this maturity and deployment-readiness. 

Beyond prediction and detection, enabling task-aware evaluation for other autonomy stack components (e.g., tracking) is another strong future direction as it would enable for the parallel co-design of autonomy modules with increased confidence in their integrated performance. Finally, since both \cref{eqn:toy_cost} and \cref{eqn:nuScenes_cost} are differentiable, an interesting area of future research is determining how downstream tasks influence detection and prediction network training.

\clearpage
\newpage

\bibliographystyle{IEEEtran}
\bibliography{ASL_papers,main}

\end{document}